\documentclass{article}
\usepackage{spconf,amsmath,graphicx}
\usepackage{amsmath}
\usepackage[table]{xcolor}
\usepackage{xcolor}
\usepackage{soul}
\usepackage{tabularx}
\usepackage{amssymb}
\usepackage{amsfonts}
\usepackage{enumitem}
\setlist{nosep, leftmargin=14pt}

\usepackage{mwe} 

\title{Deep Learning for Automated Detection of Breast Cancer in Deep Ultraviolet Fluorescence Images with Diffusion Probabilistic Model}
\name{Sepehr Salem Ghahfarokhi\textsuperscript{1}, Tyrell To\textsuperscript{2}, Julie Jorns\textsuperscript{3}, Tina Yen\textsuperscript{4}, Bing Yu\textsuperscript{5}, Dong Hye Ye\textsuperscript{1}}
\address{Department of Computer Science,
Georgia State University, Atlanta, GA, USA\textsuperscript{1},
\\
Department of Electrical and Computer Engineering, Marquette University, Milwaukee, WI, USA\textsuperscript{2},
\\
Department of Pathology, Medical College of Wisconsin, Milwaukee, WI, USA\textsuperscript{3},
\\
Department of Surgery, Medical College of Wisconsin, Milwaukee, WI, USA\textsuperscript{4},
\\
Department of Biomedical Engineering, Marquette University, Milwaukee, WI, USA\textsuperscript{5}
}

\begin{document}
%
\maketitle
\begin{abstract}
Data limitation is a significant challenge in applying deep learning to medical images. Recently, the diffusion probabilistic model (DPM) has shown the potential to generate high-quality images by converting Gaussian random noise into realistic images. In this paper, we apply the DPM to augment the deep ultraviolet fluorescence (DUV) image dataset with an aim to improve breast cancer classification for intra-operative margin assessment. For classification, we divide the whole surface DUV image into small patches and extract convolutional features for each patch by utilizing the pre-trained ResNet. Then, we feed them into an XGBoost classifier for patch-level decisions and then fuse them with a regional importance map computed by Grad-CAM++ for whole surface-level prediction. {Our experimental results show that augmenting the training dataset with the DPM significantly improves breast cancer detection performance in DUV images, increasing accuracy from 93\% to 97\%, compared to using Affine transformations and ProGAN.}
\end{abstract}
\section{Introduction}
\label{sec:intro}
Deep ultraviolet fluorescence scanning microscopy (DUV-FSM) provides rapid whole-surface imaging of dissected tissue during breast-conserving surgery without the need for invasive techniques or excessive sectioning. DUV images are particularly helpful in identifying cancer cells at the surgical specimen's edge (margin), thanks to their clear color and texture differences from healthy tissue. Then, an automated breast cancer detection in DUV images is required for intra-operative margin assessment. Deep learning-based methods have shown potential in medical image classification, but they often face due to their reliance on extensive training data~\cite{lu2020rapid},\cite{lu2022automated}. This challenge is especially notable in the classification of DUV images with a limited number of subjects, given its novelty~\cite{lan2020generative}. 

Data augmentation techniques are used to boost the medical image dataset. A widely used augmentation technique is Generative Adversarial Networks (GAN) introduced by Goodfellow et. al.\cite{goodfellow2020generative}, with its later variants being the most common models for creating synthetic images. For example, SH Gheshlaghi et. al.\cite{gheshlaghi2021breast} employed an Auxiliary Classifier Generative Adversarial Network (ACGAN) to augment a small dataset with realistic images and class labels, specifically focusing on breast cancer histopathological image classification. 
Nevertheless, GANs tend to capture a lower degree of diversity in generated content when compared to contemporary likelihood-based models \cite{razavi2019generating},\cite{nichol2021improved},\cite{nash2021generating}. Additionally, GANs can be challenging to train, often susceptible to issues such as mode collapse, which can be mitigated through meticulous hyperparameter selection and the application of suitable regularizers ~\cite{brock2018large},\cite{miyato2018spectral}.

To tackle these challenges, we apply the diffusion probabilistic model (DPM) in data augmentation to generate realistic and diverse DUV images. DPM has recently surfaced as a potent generative model, positioning itself as a potential substitute for GANs \cite{dhariwal2021diffusion}. The DPM harnesses cross-attention and adaptable conditioning to facilitate the creation of desired images. Commencing with Ho et. al.\cite{ho2020denoising}, a series of studies have demonstrated that DPMs have the ability to produce high-fidelity images akin to those produced by GANs \cite{nichol2021improved},\cite{choi2022perception}. These models offer a range of advantageous qualities for image synthesis, including stable training. \cite{moghadam2023morphology} utilized a DPM for the synthesis of histopathology images and compared it with ProGAN. Generative metrics demonstrated the superiority of the diffusion model with respect to data augmentation~\cite{zhou2022u},{\cite{kim2023adaptive}. In this paper, we adopt the DPM to enhance deep learning-based breast cancer detection in DUV images. 
The key contributions of this paper can be summarized as follows:
\begin{itemize}
    \item DPMs are employed to generate authentic DUV images, representing the initial use of this application.
    \item DUV breast cancer detection is enhanced through DPM's augmented training, surpassing GAN performance.
\end{itemize}
\begin{figure*}[h!]
  \includegraphics[width=\textwidth]{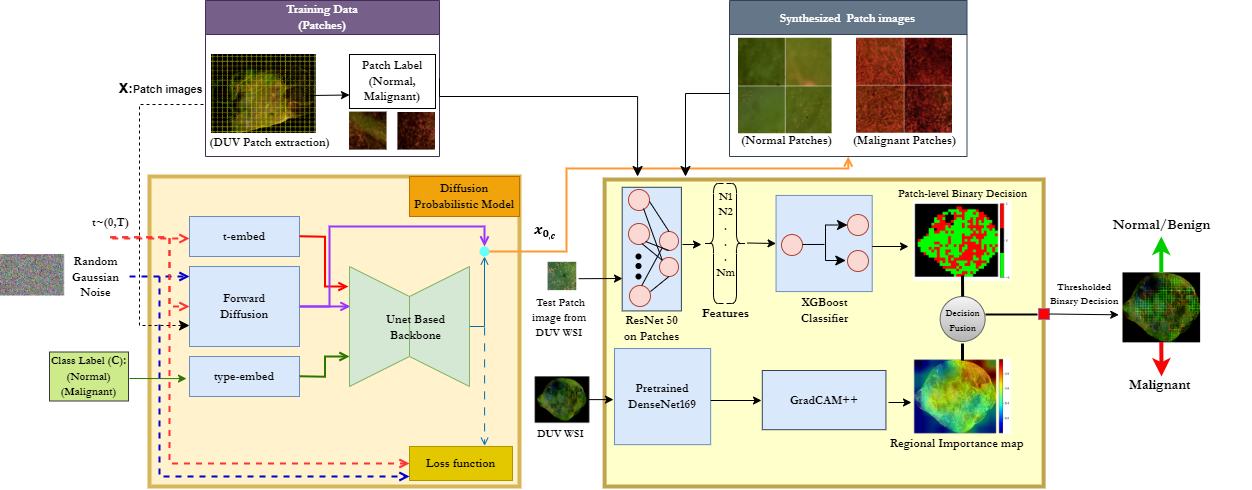}
  \caption{ Overview of the proposed method:
The proposed method starts by extracting patches from Whole Surface Images (WSI-DUV). A two-step diffusion process, involving noise addition and removal with probabilistic models, is applied to generate patch images. Utilizing a generated dataset alongside an existing training dataset, deep convolutional features are extracted using a pre-trained ResNet50 network. Patch-level classification is then performed using XGBoost, and a regional importance map is computed with Grad-CAM++ on a pre-trained DenseNet169 model for DUV-WSI. The final prediction at the WSI level is achieved through a decision fusion approach, combining patch-level results with the regional importance map. } 
\end{figure*} 
\section{METHOD}
\label{sec:pagestyle}
This section outlines our utilization of the DPM in the context of breast cancer classification in DUV images, as described in Figure 1. The pivotal role of the DPM lies in its capacity to generate synthetic DUV patch images, addressing the challenge of limited training data. Subsequently, the augmented dataset is employed to extract convolutional features using a pre-trained ResNet50 network. Ultimately, the patch-level classification is executed with the aid of the XGBoost classifier and fused into a whole-surface-level decision with a regional importance map. The ensuing sections will delve into each of these steps in greater detail.
\subsection{Diffusion Probabilistic Model}
\label{sec:majhead}
Diffusion probabilistic models (DPM) fall under the category of generative models, aiming to produce data resembling their original training data. These models function by iteratively introducing random noise to the training data (forward diffusion) and subsequently learning to eliminate this noise (reverse diffusion). Once trained, the DPM can generate new data by applying random noise through the learned process that eliminates the noise. We apply DPM to augment the training dataset by considering DUV patch images as the input, denoted as $x$. Specifically, a DUV WSI for a sample $i$ is divided into multiple DUV patches where each sample's field of view $\Omega_i$ is the union of non-overlapping patches $\Omega_i^j$ such that $\Omega_i = \cup_{j=1}^{N}\Omega_i^j, \text{and } \Omega_i^k\cap\Omega_i^{l}=\emptyset \text{ for } \forall{k,l}$.  

The DPM includes two steps: forward and reverse diffusion. It's important to note that our patch images are categorized into two distinct class labels (type-embed) represented by $c$, which $c\in\left \{Benign, Malignant\right \}$. In the forward diffusion step, we add random Gaussian noise to our data repeatedly, for a certain number of times called $T$ (t-embed) until it reaches the desired complex data points distribution. If we label the data distribution for our input as $q(x_{0_{,c}})$, the forward process defined as following steps:
 \begin{equation}
     q(x_{t_{,c}}|x_{t-1_{,c}})=\mathcal{N}(x_{t-1_{,c}}\sqrt{1-\beta_{t}},I\beta_{t}),
 \end{equation}
 where $\beta_{t}\in(0,1)$ is noise scales. By using reparametrization $x_{t}$ can be expressed as a linear combination of $x_{0}$ and a Gaussian noise variable $\varepsilon=\mathcal{N}(0,I)$:
 \begin{equation}
x_{t_{,c}} = \sqrt{\alpha_t}x_{0_{,c}} +\sqrt{1 - \alpha_t} \epsilon,
 \end{equation}
\begin{equation}
   \alpha_{t}=\prod_ {t=1}^{T}1-\beta_{_{t}}.
\end{equation} 

For the reverse diffusion step, we want to generate a sample from $q(x_{t-1_{,c}}|x_{t_{,c}})$. Since $q(x_{t-1_{,c}}|x_{t_{,c}})$ is an unknown distribution, we train a neural network $p_{\theta }(x_{t-1_{,c}}|x_{t_{,c}},\alpha_{t})$ to approximate it. To generate a random sample in the reverse diffusion, the latent variable $x_{t{,c}}$ should approximately follow an isotropic Gaussian distribution. This implies that key variables, including $\alpha_{t}$, need to be very close to zero, and $\beta_{t}$ should also have a small value, ensuring that $x_{t_{,c}}\sim\mathcal{N}(0,I)$. The network for $p_{\theta}$ has a similar role to the decoder network in variational
autoencoder (VAE). Notably, the encoder in the DPM differs from VAE in that it constitutes a fixed forward diffusion process.
Within the reverse diffusion, a neural network $\varepsilon_{\theta}$ with parameters $\theta$ learns to denoise the provided $x_{t_{,c}}$, producing $x_{t-1_{,c}}$ as output. This denoising process involves iteratively subtracting the predicted noise from the neural network.
We utilized a U-Net neural network architecture with ResNet Blocks as its backbone, as introduced by \cite{he2016deep}. This architectural choice was implemented in both the downsampling and upsampling blocks of the U-Net which has 23 convolutional layers and two residual blocks (1000 denoising diffusion steps and a learning rate of $1e-4$). The loss function for the DPM guides the model to generate synthesis images closely matching the desired distribution. Comprising two main parts, $L_{simple}$ and $L_{vlb}$, the loss function combines to minimize the difference between actual and estimated noise in the generated images\cite{vincent2011connection}:
\begin{equation}
    Loss= L_{simple}+L_{vlb},
\end{equation}
where $L_{simple}$ focuses on the difference between the actual and estimated noise in the synthesis images and it involves a mean-squared error and $L_{vlb}$ is a sum of score-matching losses and helps in learning the standard deviation $\sigma _{t}z$ during the diffusion process.
The final generated image $x_{0_{,c}}$ at the end of these iterations is expressed by the following:
\begin{equation}
    x_{0_{,c}}=\frac{1}{\sqrt{1-\beta_{t}}}(x_{t_{,c}}-\beta_{t}\sqrt{1-\alpha_{t}}\varepsilon _{\theta}(x_{t_{,c}},t))+\sigma _{t}z,
\end{equation}
where $\sigma _{t}z$ represents the noise added to the generated data at a particular diffusion step or time step $t$.

\subsection{Deep learning classification of DUV images}
Given generated patch DUV images in two labels, represented by $x_{0_{,c}}$, we add them to our training data to improve our breast cancer detection. Employing the deep learning-based breast cancer classification method for DUV images~\cite{to2023deep}, $N$ DUV patches, consisting of both generated and original patch DUV images (denoted as $p_{i}^{j},j=\left \{ 1,..., N \right \}$), are categorized benign ($-1$) or malignant ($+1$). Features are extracted using the final layer of a pre-trained ResNet50~\cite{he2016deep}, and an XGBoost classifier~\cite{shokouhmand2021efficient} assigns a binary output $y_{i}^{j}\in \left \{-1,+1 \right \}$ to each patch $p_{i}^{j}$. Additionally, Grad-CAM++\cite{selvaraju2017grad} on the pre-trained DenseNet169 model calculates the regional importance map $r_{i}^{j}$ for each DUV patch by taking the average relevance value over a patch's region $\Omega _{i}^{j}$~\cite{to2023deep}. Finally, a decision fusion method is employed to determine the WSI-level classification label $L_{i}\in \left \{-1,+1 \right \}$ based on the patch-level classification labels $y_{i}^{j}$ for all patches $j=\left \{ 1,...,m \right \}$. Toward this, we define the weight $w_i^j$ for each patch $p_i^j$ as the thresholded regional importance value $r_i^j$.
\begin{equation}
    w_i^j = \begin{cases}
    0 & \text{if } r_i^j < 0.25 \\
    r_i^j & \text{otherwise}
    \end{cases}
\end{equation}
This weighting scheme neglects patches with low importance for either malignant or benign conditions in the fused decision for the DUV WSI. Then, the weighted majority voting is employed to determine the WSI-level classification label $L_i\in\{-1,+1\}$.
\begin{equation}\label{eq:3}
L_i = sign(\sum_{j=1}^{m} w_i^{j} \cdot y_i^j), 
\end{equation}
where $sign(\cdot)$ is the sign function to map positive (malignant) and negative (benign) values to $-1$ and $+1$, respectively. It is worth noting that the DPM-augmented dataset mitigates the risk of overfitting. 

\begin{figure*}[h!]
   \centering
  \includegraphics[width=1\textwidth]{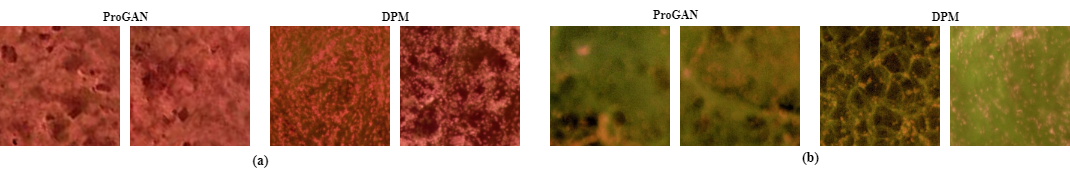}
  \caption{Comparison of synthesized patch images using ProGAN and DPM: The images generated by DPM closely mimic real biological features, showcasing characteristics like enlarged cells, dense cellular structures, infiltration, and varied nuclear traits in both malignant and benign types, in contrast to ProGAN (a: malignant, b: benign).it's apparent that DPM produces high-quality, sharp data samples with intricate features, while ProGAN generates images with noticeable blurring artifacts. Additionally, our method stands out due to its capability to capture a diverse range of image variations.}
\end{figure*}

\section{Experiment}
\label{sec:majhead}
In this study, we employed the Diffusion Probabilistic Model (DPM) to enhance our breast cancer DUV image classification. The training dataset is augmented with 1000 synthesized patches, evenly distributed between benign and malignant labels. These images, along with disease labels, seamlessly augment the training dataset without additional input from pathologists, addressing the limitations of small datasets. For comparison, we used the traditional affine transform (rotation and flip) and ProGAN\cite{karras2017progressive} to boost the DUV patch training dataset. It is important to mention that we trained separate ProGAN networks for the benign-only and malignant-only datasets to ensure the automatic assignment of labels to generated images, as ProGAN does not incorporate label information. We evaluated the results in two parts: Visual Inspection and Classification Performance. 
\subsection{Dataset}
\label{sec:majhead}
The breast cancer dataset consists of DUV images from 60 samples (24 normal/benign and 36 malignant). This DUV dataset was collected from the Medical College of Wisconsin (MCW) tissue bank (4) with a custom DUV-FSM system. The DUV-FSM used a deep ultraviolet (DUV) excitation at 285 nm and a low magnification objective (4X), which achieved a small spatial resolution from 2 to 3 mm. To enhance fluorescence contrast, breast tissues are stained with propidium iodide and eosin Y. This technique produces images of the microscopic resolution, sharpness, and contrast from fresh tissue stained with multiple fluorescence dyes. Following the extraction of patches from these images, the dataset comprises 25,024 patches from normal/benign cases and 9,444 patches from malignant cases for training our diffusion model.

\subsection{Visual Inspection}
Our examination involves visual comparison of the DUV patches generated through DPM with those produced by the proposed ProGAN, as depicted in Figure 2. 
DPM excels in generating high-quality, sharply detailed data samples with intricate features that closely replicate real biological characteristics. These characteristics include enlarged cells, dense cellular structures, infiltration, and diverse nuclear traits in both malignant and benign types. In contrast, ProGAN exhibits noticeable blurring artifacts in its generated images. Furthermore, our method demonstrates its capability to handle a broad spectrum of diverse image styles. For example, in malignant patch images, a color combination ranging from red to light green is achieved, while in benign patch images, a graceful shift to green tones is observed. Conversely, the two ProGAN networks with organized data do not produce a wide range of image diversity as part of their output.
\label{sec:majhead}

\subsection{Classification Performance}
\label{sec:majhead}
For a quantitative assessment, we gauged the classification performance of our method by juxtaposing it with the Affine Transform and ProGAN approaches when integrating synthesized images from each into our original dataset. Table 1 presents the outcomes of the 5-fold cross-validation for classification performance, underscoring the efficacy of our proposed method (DPM). It substantially enhanced accuracy to (97\%), achieving noteworthy sensitivity (97\%) and specificity (93\%). In contrast, ProGAN failed to enhance classification performance compared to the value of Affine Transform about (93\%) of accuracy. Importantly, it should be noted that while WSI-level accuracy remains consistent, there are variations in the patch-level classification results for Affine Transform and ProGAN. These findings underscore the resilience of our proposed method, in stark contrast to the over-fitting issues faced by the Affine Transform and ProGAN. Noteworthy is the observation of remarkably high sensitivity and specificity, highlighting the benefits of our approach in intra-operative margin assessment. This suggests its potential to significantly mitigate the risk of cancer recurrence by minimizing the probability of surgeons misidentifying breast cancer margins in dissected tissue. {Considering that separate ProGAN networks were trained for benign-only and malignant-only datasets due to its inability to directly incorporate label information, this approach illustrates ProGAN's limitations in generating labeled, class-specific images. In contrast, DPM's flexibility and effectiveness in handling complex data distributions and producing accurately labeled images offer a solution to the challenges of limited data sizes and suboptimal training conditions faced by ProGAN.}
\begin{table}
\centering
\caption{Results from ten 5-fold cross validations (means and standard deviations) with randomized seeds:}
\begin{tabular}{|c|c|c|c|}
    \hline
    \textbf{} & \textbf{(1) Affine} & \textbf{(2) ProGAN} & \textbf{(3) DPM} \\
    \hline
    Accuracy & $93\% \pm 0.69$
 & $93\% \pm 0.74$
 & $97\% \pm 1.23$
\\
    \hline
    Sensitivity & $94\% \pm 0.58$
 & $94\% \pm 0.61$ & $97\% \pm 1.13$
 \\
    \hline
    Specificity & $76\% \pm 4.3$
 & $76\% \pm 5.25$
 & $93\% \pm 7$
 \\
    \hline
\end{tabular}
\end{table}
\section{CONCLUSION}
\label{sec:pagestyle}
This study presents a compelling solution to the challenge of limited data in deep learning applications for medical image analysis, particularly in breast cancer classification of DUV images. Utilizing a Diffusion Probabilistic Model, the research effectively generates synthetic DUV patch images, thereby augmenting the dataset and improving the performance of deep learning models. The approach emphasizes diverse morphology levels during image synthesis, resulting in a dataset of 1000 patch images that encompasses both benign and malignant cases. Quantitative results demonstrate a significant enhancement in performance, with accuracy, sensitivity, and specificity reaching 97\%, 97\%, and 93\%, respectively. This highlights the approach's potential to greatly improve breast cancer detection. The study underscores the efficacy of data augmentation techniques in addressing data limitations in medical image analysis, ultimately contributing to more accurate and robust diagnostic systems. {However, due to the unique and original nature of our dataset, which requires extensive review by pathologists for accurate labeling, future work will focus on expanding our dataset with more labeled images to enhance our model's performance.}
\bibliographystyle{IEEEbib}
\bibliography{strings,refs}

\end{document}